\documentclass[11pt,letterpaper]{article}
\usepackage{emnlp2017}
\usepackage{times}
\usepackage{latexsym}
\usepackage{graphicx}

\emnlpfinalcopy

\def\emnlppaperid{***}

\title{A Syntactic Approach to Domain-Specific Automatic Question Generation}

\author{Guy Danon \\
	Department of Software and \\Information Systems Engineering \\
	Ben-Gurion University of the Negev \\
	Beer-Sheva 84105, Israel \\
	{\tt guy.danon@gmail.com} \\\And
	Mark Last \\
	Department of Software and \\Information Systems Engineering \\
	Ben-Gurion University of the Negev \\
	Beer-Sheva 84105, Israel \\
	{\tt mlast@bgu.ac.il} \\}

\date{}

\begin{document}
	
	\maketitle
	
	\begin{abstract}
		Factoid questions are questions that require short fact-based answers. Automatic generation (AQG) of factoid questions from a given text can contribute to educational activities, interactive question answering systems, search engines, and other applications.  
		The goal of our research is to generate factoid source-question-answer triplets based on a specific domain.  We propose a four-component pipeline, which obtains as input a training corpus of domain-specific documents, along with a set of declarative sentences from the same domain, and generates as output a set of factoid questions that refer to the source sentences but are slightly different from them, so that a question-answering system or a person can be asked a question that requires a deeper understanding and knowledge than a simple word-matching. Contrary to existing domain-specific AQG systems that utilize the template-based approach to question generation, we propose to transform each source sentence into a set of questions by applying a series of domain--independent rules (a syntactic-based approach).
		Our pipeline was evaluated in the domain of cyber security using a series of experiments on each component of the pipeline separately and on the end-to-end system. The proposed approach generated a higher percentage of acceptable questions than a prior state-of-the-art AQG system.
	\end{abstract}

\section{Introduction}

Automatic Question Generation (AQG) is defined by Rus et al.~\shortcite{rus2008question} as “the task of automatically generating questions from various inputs such as raw text, database, or semantic representation”. Question generation is an important element of learning and educational environments, search engines, automated help systems, and other applications. Manual writing of good questions, though, is a challenging and time-consuming task. The increasing availability of electronic information, along with the growth of various question--answering applications, has stimulated the research in automatic question generation.


Studies on automatic question generation regard the question generation task as a four--stage process: 1) defining the trigger to ask the question (relevant mostly to dialogs and question-answering systems), 2) text preprocessing and relevant content selection, 3) question construction, 4) ranking the constructed questions.


The main step in the AQG process is the construction of questions from the selected content. Nearly all existing AQG systems generate questions from one sentence at a time.  Existing question generation approaches can be classified into three categories: syntax--based, template--based and semantics--based~\cite{yao2010question}. Heilman~\shortcite{heilman2011automatic} presents a syntax--based system, which breaks the AQG process into several steps. Simplified factual statements are first extracted from complex inputs, by (optionally) altering or transforming lexical items, syntactic structure, and semantics. Next, the sentences are separately transformed into questions by applying sequences of simple, linguistically--motivated transformations such as subject--auxiliary inversion and WH-movement. Heilman employs some core NLP tools in his system in order to analyze the linguistic properties of input sentences.

The template-based approaches rely on the idea that a question template can capture a class of context specific questions having a common structure. For example, Chen et al.~\shortcite{chen2009aist}  developed templates such as: "What would happen if $ <X> $?" for conditional text, and "Why $ <auxiliary-verb><X> $?" for linguistic modality, where $ <X> $ is the place-holder mapped to semantic roles annotated by a semantic role labeler. In a more recent paper, Serban et al.~\shortcite{serban-EtAl:2016:P16-1} generate a 30M factoid question--answer corpus, where subject--relationship--object triples are transduced into questions about the subjects with the objects being the correct answer. Such question templates can only be used for specific entity relationships. These approaches are most suitable for special--purpose applications within a closed domain. For example, Lee et al.~\shortcite{DBLP:journals/corr/LeeKYJKY16} use a small portion of the above 30M question--answer pairs for training an academic instance of the IBM Watson system for the domain of location--related questions.

Each of the different AQG approaches has its own advantages. While the template-based approaches usually generate questions that are grammatically correct, the syntax-based approaches provide better coverage of the text, and the semantic-based approaches use some background resources, such as WordNet and Wikipedia, in order to provide more challenging questions. Some of the recent studies try to combine the different approaches.  Mazidi and Nielsen~\shortcite{mazidi2015leveraging} present a template-based question generation system that is built on multiple views of text: syntactic structure retrieved from dependency parsing, coupled with information from semantic role labels and discourse cues. 

Using various AQG approaches, an exhaustive list of questions can be generated from each text. High percentage of these questions might be unacceptable, due to incorrect syntactic structures, non-relevance to the main topics in the text or other various reasons. It is time--consuming to check the generated questions manually. Therefore in the AQG task, high precision rates are much more important than high recall~\cite{afzal2014automatic} and it can be worthwhile to apply a ranking method to the list of candidate questions. A statistical model of question acceptability, which is based on least squares linear regression, is proposed by Heilman~\shortcite{heilman2011automatic}. He models the question quality as a linear function from a vector of feature values to a continuous variable ranging from 1 to 5, representing linguistic factors such as grammaticality, vagueness, use of an appropriate WH word, etc. In total, his question ranker uses $ 179 $ features that were identified by an analysis of questions generated from development data.

Most AQG systems build wh--questions by applying some transformation rules to the original sentence. One drawback of these systems is that generated questions may be too close to the source text, as generation usually relies on transforming the syntactic structure from declarative to interrogative, without changing the words used. Such questions can be easily answered using the words in the source text even without a good comprehension of the topic. To generate more challenging questions, the source text should be paraphrased in the process of question generation. 
In the context of AQG, paraphrasing was used in only a few studies. Bernhard et al.~\shortcite{bernhard2012question} describe an AQG system for French, which reformulates questions based on variations in the question words, inducing answers with different granularities, and nominalizations of action verbs. Tseng et al.~\shortcite{tseng2014generating} generate multiple-choice questions for reading comprehension tests.  Their Paraphrase Generation System generates a ranked list of paraphrases given an input sentence and a source article, by combining multiple paraphrase resources.

Our research is aimed at generating a diverse set of factoid source-question-answer triplets representing a specific domain (e.g., health care or cyber security). These triplets can be used for both training and testing an advanced QA system such as IBM Watson. Since such a system is expected to answer natural language questions, its training set should not be limited to questions having a specific structure, which are usually produced by template--based AQG systems.  Precision of the automatically generated questions is less important than the recall of the domain knowledge, since manual filtering of these questions by human experts is much less labor--intensive than composing new questions from scratch.  Using their background knowledge, humans may ask a question about the same fact in many different forms.  To provide a variety of human--like questions, our AQG system should be able to paraphrase the source texts before transducing them into questions.  Too general questions, like "Who was born in Hawaii?", should be automatically replaced with more specific questions (e.g., "Which US President was born in Hawaii?").  We extend a state--of--the--art syntax--based AQG system~\cite{heilman2011automatic} to deal with the above challenges and evaluate our approach on questions from the cyber security domain.

\section{Proposed AQG Methodology}

Our pipeline for automatic question generation consists of four components, which together provide solutions to some of the gaps that we have found between the human generated questions and the questions which were generated using Heilman's~\shortcite{heilman2011automatic} AQG system. First we train a Word2Vec model~\cite{mikolov2013efficient} based on a domain specific corpus, after lemmatization and part-of-speech tagging.  The input to the question generation process is a set of sentences related to the domain on which the Word2Vec model was trained.  In the paraphrasing component, we identify all verbs in the sentence and try to substitute them with nearly synonymous verbs, while keeping the meaning and the syntactic correctness of the sentence. The paraphrased sentence is the input to Heilman's AQG system~\cite{heilman2011automatic}, which generates wh--questions using transformation rules. The purpose of the fourth component in the pipeline is to identify the hypernym of each answer and add it to a "what type of" question.  The generated questions using Heilman's system may be ambiguous or meaningless, because it replaces the whole answer-phrase with the wh-word, which yields, for example, "Who was born in Switzerland?". We try to identify the answer's hypernym using two strategies: from background knowledge resources or from the sentence itself. If a hypernym is found, it is used to generate a more specific question. 

\begin{figure*}[!t]
	\center
	\includegraphics[height=8cm]{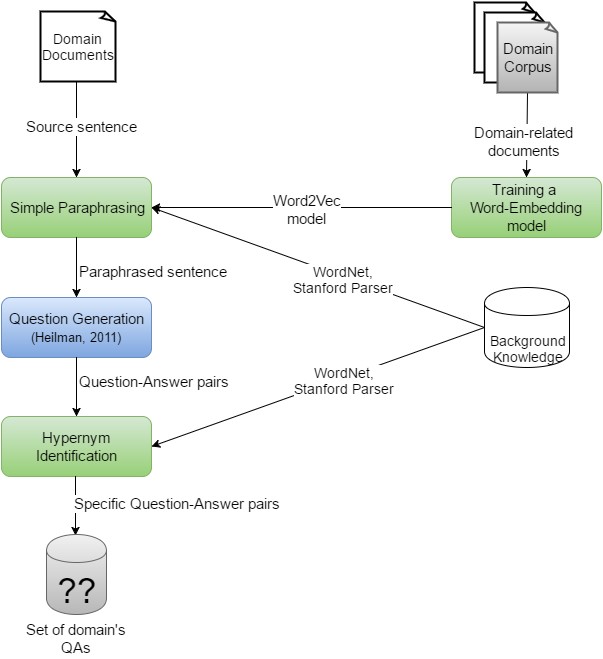}
	\caption[Pipeline for automatic factoid question generation]{AQG pipeline}
	\label{fig:pipeline}
\end{figure*}

The pipeline is shown in~\ref{fig:pipeline}. The green components are our extensions to the baseline state-of-art AQG system~\cite{DBLP:journals/corr/abs-1301-3781}. Using these extensions we would like to generate better factoid questions from a domain specific corpus.
A detailed description of each extension is provided below.

\subsection{Training a Domain--Specific Word Embedding Model}

Word embedding models embed an entire vocabulary into a relatively low-dimensional linear space, whose dimensions are latent continuous features. The embedded word vectors are trained over large collections of text using variants of neural networks, in order to capture attributional similarities between vocabulary items: words that appear in similar contexts will be close to each other in the projected space~\cite{levy2014neural}. The extraordinary abilities of this representation allow to capture valuable semantic information, which is difficult to find in other sources. For example: in WordNet the synonyms of ‘install’ are ‘instal’, ‘put in’, ‘set up’, and ‘establish’, but using a domain-specific embedding model some other synonymous terms can be found (such as ‘run’ or ‘execute’, which are relevant to the computing domain).  
The word embedding models can contribute to the AQG task in two ways: (1) There is no need to spend the expensive time of domain experts on representing all domain--related entities and their relations, and (2) we can use these models to find similarities and relations between words in order to create some paraphrases by word substitutions.

We trained the Word2Vec model~\cite{mikolov2013efficient} over a corpus of more than 1.1 million cyber-related documents.  Before training, the corpus was lemmatized and tagged by basic POS tags to improve the quality of the paraphrasing process.  Thus, we would like to generate two different vectors for the word \textit{run}: \textit{run\_nn} (\textit{run} as NOUN) and \textit{run\_vb} (\textit{run} as VERB).  However, different forms of the verb \textit{run} (\textit{run \/, runs \/, running \/, ran}) should be represented by the same vector (\textit{run\_vb}).  The Word2Vec training parameters were adjusted using a set of domain-specific examples such as:
\begin{itemize}
	\item The similarity between \textit{virus\_nn} and \textit{malware\_nn} should be relatively high. 
	\item The most similar words of \textit{run\_vb} should include verbs such as \textit{execute\_vb}.
	\item The most similar words of \textit{virus\_nn + install\_vb} should include \textit{infect\_vb}.
	\item One of the answers to \textit{Melissa\_nn is to virus\_nn as DDOS\_nn is to ‘?’} should be \textit{attack\_nn}.
\end{itemize} 

\subsection{Verb Paraphrasing}

To generate a question, which is slightly different from the source text, we substitute the verbs in the original sentence by other verbs having a similar meaning and then submit the paraphrased sentence to a question generation system. 
We identify the verbs in each sentence by the following POS tags from Stanford Parser: ‘VB’, ‘VBD’, ‘VBG’, ‘VBN’, ‘VBP’, and ‘VBZ’, except words with the lemma \textit{be}. Adverbial particles are attached to the verbs according to the dependency tree of the sentence in order to consider them as a part of a compound verb. For example: \textit{shut down} and \textit{boost up} are compound verbs, and not just \textit{shut} or \textit{boost}. When a word in a sentence is tagged as ‘RP’ (particle), we iterate over all its dependencies and attach it to the verb with which it has the \textit{prt} (phrasal verb particle) relation.

Inspired by the studies like~\cite{bolshakov2004synonymous} that used WordNet for paraphrasing, we search the WordNet for the synonyms of all verbs in each source sentence. Since the synonyms in WordNet are grouped by senses, we retrieve all synonyms from all verb synsets of each word identified as a verb. The main drawback of WordNet is that it is a general purpose lexicon / ontology and therefore it may miss some domain-specific actions. On the other hand, since WordNet was generated by human experts, the retrieved synonyms are certainly right (high precision). 

Multiple potential synonyms may be retrieved for each verb. For example, the verb \textit{run} has $ 41 $ different senses and most of them contain several synonyms. 
Following Lee et al.~\shortcite{lee2016combining}, we use  a domain-specific Word2Vec model to disambiguate the verb senses.  The Word2Vec objective function causes words that occur in similar contexts to have similar embeddings but in different domains the context of the same word will probably be different.  For each potential synonym, we compute its similarity to the original verb from the source sentence, which is the cosine similarity between the vectors of the two words. The synonym which has the highest similarity with the source verb is selected to substitute it in the sentence, unless this similarity is below a user-defined min\_similarity threshold.

The words in the WordNet and in a domain-specific Word2Vec model appear in their base form (lemma). To substitute the original verb with the selected synonym, the synonym has first to be transformed to the form of the original verb in the sentence. We have defined a set of domain--independent rules in SimpleNLG~\cite{gatt2009simplenlg} for this purpose. The rules refer to the following verb parameters: (1) tense, (2) number (singular / plural), (3) progressive (yes / no) and (4) passive (yes / no).  

\subsection{Question Generation} 

Given the paraphrased sentence, we use a state-of-art system to generate various questions based on the factual statements in the sentence~\cite{heilman2011automatic}. As mentioned in our first principle, we use the syntactic-based approach to question generation in order to achieve good coverage (high recall) of the text, albeit at the expense of precision. The AQG system presented by Heilman~\shortcite{heilman2011automatic} has three stages: NLP transformations on the source sentences, question transducer, and question ranker. We have integrated the first two stages in our question generation pipeline. The ranking stage was not included since we assume that the automatically generated questions can be easily filtered by a human expert before being released for any practical usage.

Heilman's system uses a Java reimplementation of the supersense tagger described by Ciaramita and Altun~\shortcite{ciaramita2006broad} as part of mapping potential answer phrases to WH words. It uses features of both the current word (e.g., part of speech, capitalization features, the word itself) and the surrounding words to make predictions. They use a set of conditions to determine the WH-word, but claim that "It is worth noting that this step encodes a considerable amount of knowledge and yet still does not cover all relevant linguistic phenomena." Following their error analysis and some of our experiments, we have decided to revise the WH-words of Heilman's questions based on the NE type of the answer phrase as determined by the Stanford parser (i.e. "who" for PERSON, "where" for LOCATION, etc.).   

\subsection{Hypernym Identification}

Good factoid questions must be specific, fluent and clear. The answer phrase determines whether the question type will be a “when” / “who” / when”, etc. Syntactic-based AQG approaches usually map the answer phrase to the relevant WH-question word using techniques involving a NER tool. For example, consider taking the following sentence and generating a question for which Hillary Clinton is the answer phrase:\\
\textit{Sentence}: Hillary Clinton was the United States Secretary of State from 2009 to 2013.\\
After recognizing that the answer is a person, the following question is generated:\\
\textit{Question}: Who was the United States Secretary of State from 2009 to 2013?\\
The generated question above is considered a good factoid question, since it provides a single and unambiguous answer. However, exploring some other question-answer pairs automatically generated from texts using the same technique reveals that a lot of them might be ambiguous and even ridiculous. For example, see the following source sentence and the question generated for the answer phrase “Tennis star Roger Federer”:\\
\textit{Sentence}: Tennis star Roger Federer was born in Switzerland. \\
\textit{Question}: Who was born in Switzerland?\\
Obviously, this question is not specific enough to be used in an exam or some other learning task.  We propose to narrow the scope of factoid questions using two methods: (1) adding hypernym from knowledge resources, and (2) adding content from the source text.

\subsubsection{Adding hypernym from knowledge resources}

Given a pair of two words $ <X,Y> $, $ X $ is considered a hypernym of a word $ Y $ if native speakers accept the sentence \textit{Y is a (kind of) X}. This semantic relation organizes the meanings of words and concepts into a hierarchical structure. It is also known as a generic/specific, a taxonomic, is-a or instance-of relation. Adding the hypernym of an answer to a question can produce a “type-of” (or “what-type”) question, which should be considered a better factoid question.

Consider again the sentence about Roger Federer. Generating a question by Heilman’s algorithm for the answer “Switzerland” will lead to the following questions:\\
\textit{Question 1}: Where was tennis star Roger Federer born? (After recognizing the answer as a location).\\
\textit{Question 2}: What was tennis star Roger Federer born in?\\
As explained earlier, these are ambiguous questions, since there are some other correct answers besides Switzerland, such as the name of the hospital where he was born or even other places (bath, airplane, home, hospital…). Adding the hypernym of Switzerland (country) to the question will narrow the scope and make it a better question:\\
\textit{Question 3}: What country was tennis star Roger Federer born in?\\
For each generated question, we search the WordNet for the related hypernym of the answer phrase. If the answer phrase exists in the WordNet, its direct hypernym in the hierarchy is returned like in the following examples:\\ $ unix \rightarrow operating\_system\\ malware \rightarrow software\\ red \rightarrow chromatic\_color\\ Bill Gates \rightarrow computer\_scientist $

\subsubsection{Adding content from the source sentence}

The source sentence and the answer phrase together with its POS (part-of-speech) tags and NE (named-entity) tags in particular, can contain some valuable information, which can be utilized to enhance questions generated by Heilman. If the Heilman generated answer contains a noun phrase, we try to include a generalized noun phrase in the question using some domain--independent rules. 

For answer phrases that contain NE, the tag of the NE can be added to the generated question. For example, the sentence “Bill Gates founded Microsoft in 1975” contains the NE “Microsoft”, that does not exist in WordNet (version 3.1 database files), but it is recognized by Stanford Parser as an ORGANIZATION. Adding this tag name to the question “What did Bill Gates found in 1975?” will create the question “What organization did Bill Gates found in 1975?”, which is more specific. The tag name of the NE is added to the question only if the NE part-of-speech is noun to avoid generating, for example, questions like "What ordinal…?", when the word \textit{first} appears in the source.

If the answer phrase contains NP prior to the NE, this NP can be added to the question. For example, “Tennis star” is prior to the NE “Roger Federer” in the sentence \textit{Tennis star Roger Federer was born in Switzerland}. When generating question with the answer phrase “Tennis star Roger Federer”, the following question can be generated instead of "Who was born in Switzerland?": "What tennis star was born in Switzerland?"

For answer phrases that are not found in the lexical database and do not contain named-entities, the head noun of the phrase can add a valuable contribution to the question. The head noun is the major noun which reflects the focus of a noun phrase and it often describes the function or the property of the noun phrase. It has been presented to play an important role in classifying what-type questions~\cite{li2008classifying}. We retrieve the head nouns of the answer phrases using Stanford CoreNLP library~\cite{manning2014stanford}.  Here is an example of improving a Heilman's question using a head noun:\\
\textbf{Source sentence}: Polymorphic virus infects files with an encrypted copy of itself.\\
\textbf{Question generated by Heilman}: What infects files with an encrypted copy of itself?\\
\textbf{Answer phrase}: Polymorphic virus.\\
\textbf{Head Noun}: Virus.\\
\textbf{New question (without paraphrasing)}: What virus infects files with an encrypted copy of itself?\\
The final step in adding information to the question is changing the WH-question word. Given a question, if a hypernym has been found using one of the above methods / rules, the WH-word is replaced by “What” followed by the hypernym (e.g., "What tennis star..." instead of "Who..."). 

\section{Experimental Evaluation}

In this section, we evaluate the proposed syntactic approach to domain-specific automatic question generation on the cyber security domain.  Since our pipeline consists of four different components, we perform two types of experiments: evaluations of the individual components of the proposed system and evaluation of the end-to-end AQG system by measuring the quality of questions at the sentence-level.  The training corpus of the Word2Vec model consists of three different sources: $ 622 $ books on cyber security, $ 2800 $ academic papers on the cyber domain, and $ 250 $ cyber reports collected from the Internet between 2008-2014. The total number of sentences in the whole corpus is more than $ 4.6 $ million (after sentence splitting using Stanford Parser).  Our evaluation corpus contains $ 69 $ sentences extracted from about $ 30 $ different articles on cyber security and composed of $ 1,130 $ words.  Most sentences were used to generate more than one question. 

\subsection{Word2Vec Model Evaluation}

For training the model, we used the implementation of Word2Vec in Python, as found in Gensim by Radim Rehurek \url{https://github.com/RaRe-Technologies/gensim}. This tool provides an efficient implementation of the continuous bag-of-words and skip-gram architectures for computing vector representations of words. Lemmatization and POS tagging were done as preprocessing to the training process, using Stanford Parser.

While we used the default values for most of the training parameters (including the architecture of CBOW), two parameters were tuned: (1) window size, which is the maximum distance between the current and predicted word within a sentence, and (2) vector size, which is the dimensionality of the feature vectors. Based on~\cite{le2014automatic}, we have experimented with different combinations of the window size parameter (in the scope of 5 to 12) and the vector size (between 50 and 200). 

To evaluate the quality of a model, we built a set of $ 20 $ examples from the cyber domain and defined the expected answer of each example.  For a given model, each example is scored between $ 0 $ and $ 1 $, so the maximal model's total score is $ 20 $. The score of an example is given according to the following semantic NLP tasks: 
\begin{itemize}
	\item Does not match (8 examples) – In these examples, the model has to choose which word from the input list does not match the others. If the model provides the correct answer – the score is 1, otherwise it is 0.
	\item Similarity (7 examples) – The similarity score between two words is the cosine similarity between their two vectors. If the expectation is to have high similarity (e.g. between worm and virus) – the answer is scored by the returned similarity value. Otherwise (if we assume that there is no high similarity between the words, e.g. table and install) – the answer is scored as 1 minus the returned value.
	\item Most similar (5 examples) – In these examples, the task is to find the top-N most similar words, while positive words contribute positively towards the similarity and negative words negatively. This method computes cosine similarity between a simple mean of the projection weight vectors of the given words and the vectors for each word in the model. In our evaluation, we use the default value of N = 10. The model's score for the ''most similar'' example is the calculated similarity of the expected answer in the returned vector. If the expected answer is not one of the top 10 answers, the given score is 0. 
\end{itemize}
The model, which achieved the highest score according to the above method, was selected for use in the paraphrasing component.  The selected model was trained with the values of 8 and 100 for the parameters window size and vector size, respectively.

\subsection{Verb Paraphrasing}

As explained in the methodology section above, we perform a simple paraphrasing of the source sentence by substituting the sentence verbs with one of their synonyms.  We use the WordNet to retrieve the list of the verb synonyms and the Word2Vec model to disambiguate the verb senses and select the most appropriate verb for each substitution. By default, the senses in the WordNet are sorted in the decreasing order of their estimated frequency and hence we consider as a baseline the first synonym (sense) found in the WordNet.  

We applied the baseline approach along with the Word2Vec--based sense disambiguation method to $ 195 $ verbs in different tenses that were found in the $ 69 $ sentences of the evaluation corpus. The comparative evaluation was performed by two raters using two discretely-valued measures: the selection of the synonym for each verb in the sentence {correct synonym, wrong synonym, no substitute has been found} and the overall quality of the paraphrased sentence {acceptable, unacceptable}.  Using domain--specific word embeddings vs. the baseline, the number of wrong synonyms was decreased significantly (from $ 91 $  to $ 35 $), as a result of selecting $ 23 $ additional correct synonyms and leaving the original verb in $ 33 $ other cases, where the similarity value was below the predefined threshold of $ 0.5 $. Out of the set of $ 69 $ sentences, only $ 27 $ sentences paraphrased by the baseline were found acceptable, compared to $ 56 $ acceptable sentences produced by domain--specific word embeddings.

\subsection{Hypernym Identification}

The question transducer stage in the AQG system~\cite{heilman2011automatic} generates multiple questions from each input sentence, according to different answer-phrases that can be found in the sentence. As mentioned above, we try to identify the hypernym of each QA pair in order to build more specific questions.  All paraphrased sentences that we have built in the previous section were provided as an input to Heilman's system, which generated $ 217 $ QA pairs. We have manually analyzed these pairs and determined which answers can be generalized (a hypernym can be found) as a benchmark for evaluating the automatically generated hypernyms.  According to this evaluation, the precision of the hypernym identification component was $ 0.76 $ and the recall was $ 0.97 $. 

\begin{figure*}[!t]
	\center
	\includegraphics[height=8cm]{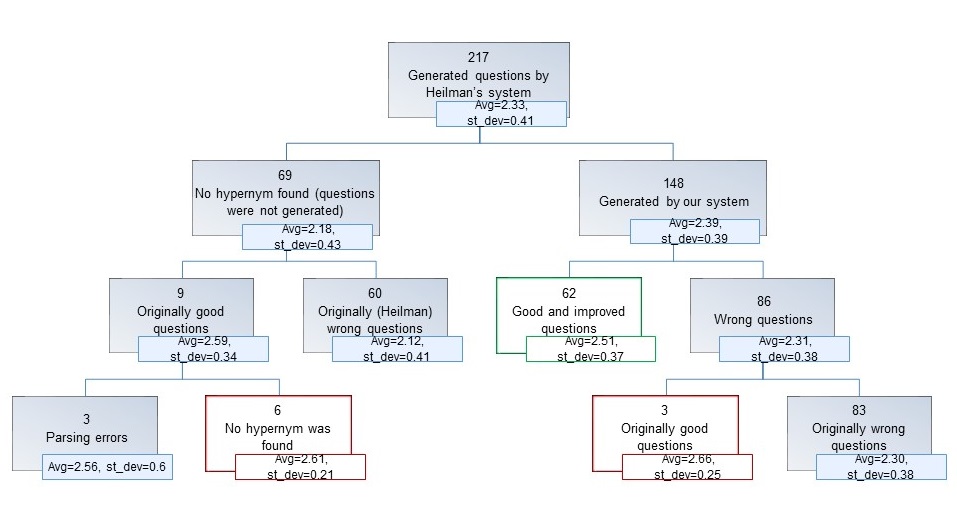}
	\caption[AQG Summary]{Summary of Experimental Results}
	\label{fig:results}
\end{figure*}

\subsection{Evaluation of the AQG Pipeline} 

We compare our questions (after verb paraphrasing and the addition of the identified hypernym of the answer) to the questions which were generated by the baseline system~\cite{heilman2011automatic}.  Two cyber security experts were asked to determine together which of the generated questions can be accepted as good questions in terms of fluency, clarity, and semantic correctness. 
In case of disagreement, there was a discussion between the two raters to reach a consensus on each generated question.
Based on this assessment, we have analyzed three different cases: (1) questions which were generated by the baseline and were not modified by our system, (2) questions which were improved by our pipeline, and (3) questions which were corrupted by our pipeline.

%

The classification of the generated questions is summarized in Fig.~\ref{fig:results}. 
The averages and the standard deviations (avg, st\_dev) in Fig.~\ref{fig:results} refer to the rankings generated by Heilman's system for the corresponding questions.  It can be seen from Fig.~\ref{fig:results} that “good” (acceptable) questions have slightly higher rankings than “wrong” (unacceptable) questions.
Out of $ 217 $ questions in total that were generated by Heilman's system, $ 148 $ questions were modified by our pipeline with an acceptance rate of 42\% (62 questions were accepted by the two rankers).
These questions, listed in the Supplementary Material, can be considered as the main contribution of our system, since they were enhanced by paraphrasing and by adding automatically detected hypernyms.
The other $ 86 $ questions were not accepted due to the following reasons:
\begin{itemize}
	\item Not interesting (trivial) QA - 26
	\item Wrong QA by Heilman	- 31
	\item Out of context - 18
	\item Wrong paraphrasing - 15
	\item Wrong hypernym - 33
\end{itemize}
Considering the fact that only three out of $ 86 $ rejected questions from our system (3.5\%) and only six out of $ 69 $ questions that were not modified by our system (8.7\%) were actually good questions, it can be concluded that we can improve the baseline QAs by generating good factoid questions that are different from the source text.

\section{Conclusions and Future Research}

In this paper, a novel syntactic-based approach for automatic question generation in a specific domain has been presented. The proposed pipeline is based on a combination of deep linguistic analysis (part of speech, named entity recognition) and knowledge resources (WordNet, Word2Vec model trained on a domain corpus) for detecting and matching entities and relations. After generating a simple paraphrased sentence by substituting the verbs of the source sentence, we use a state-of-the-art AQG system~\cite{heilman2011automatic} and add hypernyms to generate specific questions (usually in the form of \textit{What type of…}).

In a series of experiments that we have performed, it was shown that (1) the proposed method of verb paraphrasing using Word2Vec and WordNet outperforms the baseline method of~\cite{heilman2011automatic}; (2) the recall of the hypernym identification component is very high and its precision is 76\%, and (3) the proposed pipeline increases the number of acceptable QAs in the task of generating good factoid questions that are different from the source text.
Further research can improve each one of the existing pipeline components and extend it with additional components, such as content selection and question ranking. 

\section*{Acknowledgments}
This research was partially supported by IBM Cyber Security Center of Excellence (CCoE), Beer Sheva, Israel.
%
\bibliography{AQG}

\begin{thebibliography}{}
\expandafter\ifx\csname natexlab\endcsname\relax\def\natexlab#1{#1}\fi

\bibitem[{Afzal and Mitkov(2014)}]{afzal2014automatic}
Naveed Afzal and Ruslan Mitkov. 2014.
\newblock Automatic generation of multiple choice questions using
  dependency-based semantic relations.
\newblock {\em Soft Computing\/} 18(7):1269--1281.

\bibitem[{Bernhard et~al.(2012)Bernhard, De~Viron, Moriceau, Tannier
  et~al.}]{bernhard2012question}
Delphine Bernhard, Louis De~Viron, V{\'e}ronique Moriceau, Xavier Tannier,
  et~al. 2012.
\newblock Question generation for french: Collating parsers and paraphrasing
  questions.
\newblock {\em Dialogue and Discourse\/} 3(2):43--74.

\bibitem[{Bolshakov and Gelbukh(2004)}]{bolshakov2004synonymous}
Igor~A Bolshakov and Alexander Gelbukh. 2004.
\newblock Synonymous paraphrasing using wordnet and internet.
\newblock In {\em International Conference on Application of Natural Language
  to Information Systems\/}. Springer, pages 312--323.

\bibitem[{Chen et~al.(2009)Chen, Aist, and Mostow}]{chen2009aist}
W~Chen, G.~Aist, and J.~Mostow. 2009.
\newblock Generating questions automatically from informational text.
\newblock In {\em Proceedings of the 2nd Workshop on Question Generation (AIED
  2009)\/}. pages 17--24.

\bibitem[{Ciaramita and Altun(2006)}]{ciaramita2006broad}
Massimiliano Ciaramita and Yasemin Altun. 2006.
\newblock Broad-coverage sense disambiguation and information extraction with a
  supersense sequence tagger.
\newblock In {\em Proceedings of the 2006 Conference on Empirical Methods in
  Natural Language Processing\/}. Association for Computational Linguistics,
  pages 594--602.

\bibitem[{Gatt and Reiter(2009)}]{gatt2009simplenlg}
Albert Gatt and Ehud Reiter. 2009.
\newblock Simplenlg: A realisation engine for practical applications.
\newblock In {\em Proceedings of the 12th European Workshop on Natural Language
  Generation\/}. Association for Computational Linguistics, pages 90--93.

\bibitem[{Heilman(2011)}]{heilman2011automatic}
Michael Heilman. 2011.
\newblock {\em Automatic factual question generation from text\/}.
\newblock Ph.D. thesis, Carnegie Mellon University.

\bibitem[{Le et~al.(2014)Le, Kojiri, and Pinkwart}]{le2014automatic}
Nguyen-Thinh Le, Tomoko Kojiri, and Niels Pinkwart. 2014.
\newblock Automatic question generation for educational applications--the state
  of art.
\newblock In {\em Advanced Computational Methods for Knowledge Engineering\/},
  Springer, pages 325--338.

\bibitem[{Lee et~al.(2016{\natexlab{a}})Lee, Kim, Yoo, Jung, Kim, and
  Yoon}]{DBLP:journals/corr/LeeKYJKY16}
Jangho Lee, Gyuwan Kim, Jaeyoon Yoo, Changwoo Jung, Minseok Kim, and Sungroh
  Yoon. 2016{\natexlab{a}}.
\newblock \href{http://arxiv.org/abs/1611.03932}{Training {IBM} watson using
  automatically generated question-answer pairs}.
\newblock {\em CoRR\/} abs/1611.03932.
\newblock
  \href{http://arxiv.org/abs/1611.03932}{http://arxiv.org/abs/1611.03932}.

\bibitem[{Lee et~al.(2016{\natexlab{b}})Lee, Ke, Huang, and
  Chen}]{lee2016combining}
Yang-Yin Lee, Hao Ke, Hen-Hsen Huang, and Hsin-Hsi Chen. 2016{\natexlab{b}}.
\newblock Combining word embedding and lexical database for semantic
  relatedness measurement.
\newblock In {\em Proceedings of the 25th International Conference Companion on
  World Wide Web\/}. International World Wide Web Conferences Steering
  Committee, pages 73--74.

\bibitem[{Levy and Goldberg(2014)}]{levy2014neural}
Omer Levy and Yoav Goldberg. 2014.
\newblock Neural word embedding as implicit matrix factorization.
\newblock In {\em Advances in neural information processing systems\/}. pages
  2177--2185.

\bibitem[{Li et~al.(2008)Li, Zhang, Yuan, and Zhu}]{li2008classifying}
Fangtao Li, Xian Zhang, Jinhui Yuan, and Xiaoyan Zhu. 2008.
\newblock Classifying what-type questions by head noun tagging.
\newblock In {\em Proceedings of the 22nd International Conference on
  Computational Linguistics-Volume 1\/}. Association for Computational
  Linguistics, pages 481--488.

\bibitem[{Manning et~al.(2014)Manning, Surdeanu, Bauer, Finkel, Bethard, and
  McClosky}]{manning2014stanford}
Christopher~D Manning, Mihai Surdeanu, John Bauer, Jenny~Rose Finkel, Steven
  Bethard, and David McClosky. 2014.
\newblock The stanford corenlp natural language processing toolkit.
\newblock In {\em ACL (System Demonstrations)\/}. pages 55--60.

\bibitem[{Mazidi and Nielsen(2015)}]{mazidi2015leveraging}
Karen Mazidi and Rodney~D Nielsen. 2015.
\newblock Leveraging multiple views of text for automatic question generation.
\newblock In {\em International Conference on Artificial Intelligence in
  Education\/}. Springer, pages 257--266.

\bibitem[{Mikolov et~al.(2013)Mikolov, Chen, Corrado, and
  Dean}]{mikolov2013efficient}
Tomas Mikolov, Kai Chen, Greg Corrado, and Jeffrey Dean. 2013.
\newblock \href{http://arxiv.org/abs/1301.3781}{Efficient estimation of word
  representations in vector space}.
\newblock {\em CoRR\/} abs/1301.3781.
\newblock
  \href{http://arxiv.org/abs/1301.3781}{http://arxiv.org/abs/1301.3781}.

\bibitem[{Rus et~al.(2008)Rus, Cai, and Graesser}]{rus2008question}
Vasile Rus, Zhiqiang Cai, and Art Graesser. 2008.
\newblock Question generation: Example of a multi-year evaluation campaign.
\newblock {\em Proc WS on the QGSTEC\/} .

\bibitem[{Serban et~al.(2016)Serban, Garc\'{i}a-Dur\'{a}n, Gulcehre, Ahn,
  Chandar, Courville, and Bengio}]{serban-EtAl:2016:P16-1}
Iulian~Vlad Serban, Alberto Garc\'{i}a-Dur\'{a}n, Caglar Gulcehre, Sungjin Ahn,
  Sarath Chandar, Aaron Courville, and Yoshua Bengio. 2016.
\newblock \href{http://www.aclweb.org/anthology/P16-1056}{Generating factoid
  questions with recurrent neural networks: The 30m factoid question-answer
  corpus}.
\newblock In {\em Proceedings of the 54th Annual Meeting of the Association for
  Computational Linguistics (Volume 1: Long Papers)\/}. Association for
  Computational Linguistics, Berlin, Germany, pages 588--598.
\newblock
  \href{http://www.aclweb.org/anthology/P16-1056}{http://www.aclweb.org/anthology/P16-1056}.

\bibitem[{Tseng et~al.(2014)Tseng, Huang, Chen, and Sun}]{tseng2014generating}
Ya-Min Tseng, Yi-Ting Huang, Meng~Chang Chen, and Yeali~S Sun. 2014.
\newblock Generating comprehension questions using paraphrase.
\newblock In {\em Technologies and Applications of Artificial Intelligence\/},
  Springer, pages 310--321.

\bibitem[{Yao and Zhang(2010)}]{yao2010question}
Xuchen Yao and Yi~Zhang. 2010.
\newblock Question generation with minimal recursion semantics.
\newblock In {\em Proceedings of QG2010: The Third Workshop on Question
  Generation\/}. Citeseer, pages 68--75.

\end{thebibliography}
\bibliographystyle{emnlp_natbib}

\end{document}